\RequirePackage{amsthm}
\documentclass[sn-mathphys-num]{sn-jnl}

\usepackage{comment}
\usepackage{graphicx}%
\usepackage{multirow}%
\usepackage{amsmath,amssymb,amsfonts}%
\usepackage{amsthm}%
\usepackage{mathrsfs}%
\usepackage[title]{appendix}%
\usepackage{xcolor}%
\usepackage{textcomp}%
\usepackage{manyfoot}%
\usepackage{booktabs}%
\usepackage{algorithm}%
\usepackage{algorithmicx}%
\usepackage{algpseudocode}%
\usepackage{listings}%
\usepackage{cleveref}
\usepackage{subcaption}

\usepackage{todonotes}
\usepackage{xcolor}
\usepackage{etoolbox}
\newtoggle{final}
\toggletrue{final} 

\excludecomment{declaration} 

\newcommand{\pw}[1]{\iftoggle{final}{#1}{{\color{orange} #1}}}
\newcommand{\yban}[1]{\iftoggle{final}{#1}{{\color{violet} #1}}}

\newcommand{\Expect}{{\mathbb E}}

\newcommand{\eExpect}{{\hat{\mathbb E}}}

\theoremstyle{thmstyleone}%
\theoremstyle{thmstyletwo}%

\theoremstyle{thmstylethree}%

\newcommand{\St}{{\mathcal S}} 
\newcommand{\Ac}{{\mathcal A}} 
\newcommand{\T}{{T}} 
\newcommand{\Reward}{{r}} 

\newcommand{\ap}{{k}} 
\newcommand{\aps}{{\mathcal K}} 

\usepackage{review}
\setrevision{1}

\begin{document}

\title[Article Title]{\begin{center}State-Novelty Guided Action Persistence\\in Deep Reinforcement Learning
\end{center}}

\author[1]{\fnm{Jianshu} \sur{Hu}}\email{hjs1998@sjtu.edu.cn}

\author*[2]{\fnm{Paul} \sur{Weng}}\email{paul.weng@duke.edu}

\author*[1]{\fnm{Yutong} \sur{Ban}}\email{yban@sjtu.edu.cn}

\affil[1]{\orgdiv{UM-SJTU Joint Institute}, \orgname{Shanghai Jiao Tong University}, \country{China}}

\affil[2]{ \orgname{Duke Kunshan University}, \country{China}}


\abstract{

While a powerful and promising approach, deep reinforcement learning (DRL) still suffers from sample inefficiency, which can be \pw{notably} improved by resorting to more sophisticated techniques to address the exploration-exploitation dilemma.
One such technique relies on action persistence (i.e., repeating an action over multiple steps).
However, previous work exploiting action persistence either applies a fixed strategy or learns additional value functions (or policy) for selecting the repetition number. 
In this paper, we propose a novel method to dynamically adjust the action persistence based on the current exploration status of the state space. 
In such a way, our method does not require training of additional value functions or policy. 
Moreover, the use of a smooth scheduling of the repeat probability allows a more effective balance between exploration and exploitation.
Furthermore, our method can be seamlessly integrated into various basic exploration strategies to incorporate temporal persistence. 
Finally, extensive experiments on different DMControl tasks demonstrate that our state-novelty guided action persistence method significantly improves the sample efficiency.
}




\maketitle

\section{Introduction}\label{Introduction}
\begin{figure}[t!]
    \centering
    \includegraphics[width=0.85\linewidth]{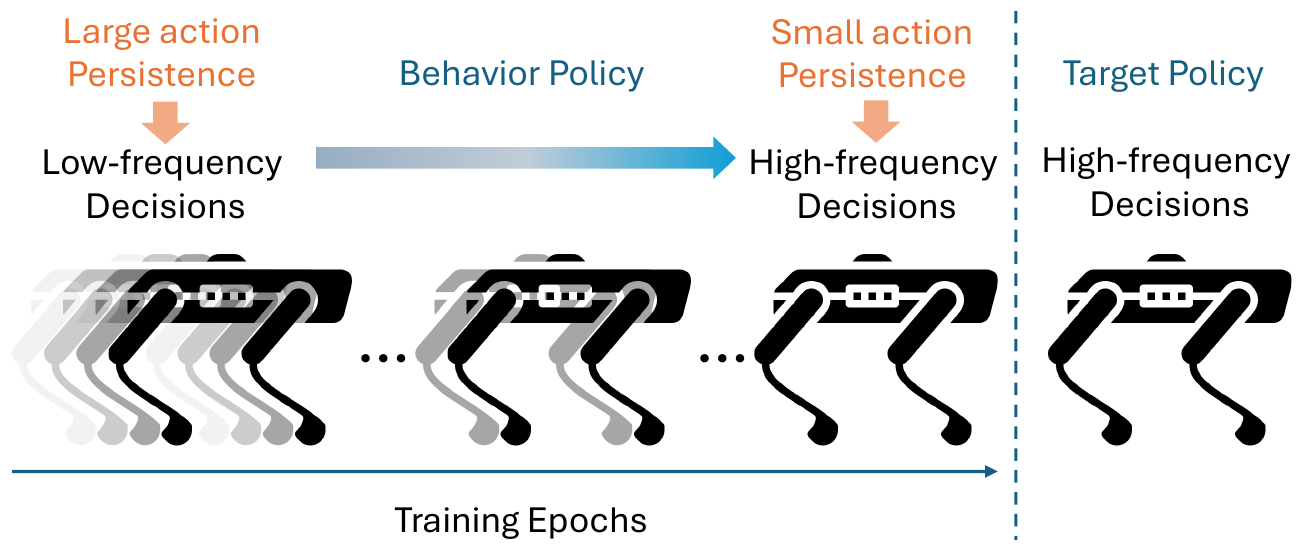}
    \caption{ 
    SNAP incorporates dynamic action persistences into the behavior policy of an off-policy DRL algorithm.
    In the initial stage, behavior policies with large action persistences make low-frequency decisions, which ensure temporally persistent exploration.
    Accordingly, fine-grained policies with small action persistences allow high-frequency actions, which is necessary for superior overall performance.} 
    \label{fig:teaser1}
    \vspace{-3ex}
\end{figure}

Deep Reinforcement Learning \pw{(DRL)} has emerged as a powerful tool for solving \pw{sequential} decision-making problems, by leveraging the impressive approximation capability of deep neural networks. 
From mastering video games \citep{atari} to solving robotics tasks \citep{learning_dexterous_manipulation} and even creating intelligent conversational agent \citep{openai2024gpt4}, DRL has demonstrated remarkable potential.
Despite the above successes, there are still some remaining issues, notably in terms of sample efficiency and the dilemma between exploration and exploitation, hindering broader applications of DRL in solving real-world problems.

In the realm of discrete-time Markov Decision Process (MDP), action persistence, also known as action repeat, is commonly used as a hyperparameter.
A properly chosen persistence can lead to a significant performance gain on the sample efficiency \citep{frame_skip_is_powerful}.
Intuitively, a larger action persistence can lead to temporally persistent exploration, 
while a smaller action persistence usually yields a higher return achieved by a fine-grained policy.
However, the optimal action persistence highly depends on the environments we are dealing with.
Existing works focus on how to dynamically choose action persistence by learning a value function or training a separate policy. These methods inevitably increase the computational complexity,  which is not preferred due to efficiency.
A simple and effective way to leverage action persistence for exploration is $\epsilon z$-greedy, which temporally extends $\pw{\epsilon}$-greedy to repeat the sampled action for a random duration.
Using a fixed distribution for action persistence is not always promising. 
For example, subtle and frequent actions are necessary in contact-rich locomotion tasks.
In this kind of environments, using a fixed distribution could result in a sub-optimal policy.

In this paper, we propose State-Novelty guided Action Persistence (SNAP) in DRL, a novel approach to dynamically deciding action repetition, guided by a state-novelty measure. 
An overview of our method is shown in \Cref{fig:teaser1}.
By evaluating the novelty of each state, our method enables the agent to adapt its action repetition strategy smoothly, striking a delicate balance between exploration and exploitation.
Moreover, without the need of training additional value functions or policies for deciding action repetition, our method circumvents significant computational overhead. 
Meanwhile, our method can be integrated into basic exploration strategies to incorporate temporally persistent exploration into them. The contribution of this paper can be summarized as follows:

\begin{enumerate}
    \item We propose a novel framework for using adaptive action persistences in the behavior policy of off-policy DRL algorithms.
    \item We carefully design an adaptive action persistence module, facilitating effective exploration by using more action repetitions in the early learning stages while seamlessly transitioning to optimal performance later by using fewer action repetitions.
    \item We experimentally demonstrate the sample efficiency of our method in DMControl tasks and show the effectiveness of temporally extending some basic exploration strategies with our method.
\end{enumerate}

\section{Related Work}\label{Related Work}

\pw{We discuss the most related work to ours and organize it into three clusters: Sample Efficient DRL, Action Persistence, and Exploration in DRL.}

\subsubsection*{Sample Efficient DRL}
Sample efficiency is crucial in DRL which motivates extensive research efforts. 
Key approaches to enhancing sample efficiency include training world models \citep{dreamerv3}, exploiting data augmentation \citep{revisiting_data_aug}, adding self-supervised loss \citep{spr} and increasing the replay ratio \citep{replay_ratio}.
Here, we give a brief introduction to Data-regularized Q (DrQ) \citep{DrQ} and DrQv2 \citep{DrQv2}, which \pw{serve} as our base algorithm due to its sample efficiency.
DrQ leverages data augmentation, specifically by applying random shift on the training samples in image-based RL tasks.
\citet{DrQv2} \pw{improve} the performance in image-based DMControl \citep{dm_control} tasks upon DrQ by introducing several changes.
One of the key changes is to switch the base DRL algorithm from SAC \citep{SAC} to DDPG \citep{DDPG}.

\subsubsection*{Action \pw{P}ersistence}
Action persistence or action repeat \citep{frame_skip_is_powerful} is commonly used in DRL to reduce the computational burden of intensively choosing action.
This technique also encourages exploration and helps with credit assignment for long-horizon tasks.
Instead of simply treating it as a hyperparameter,
research work has been conducted to decide action persistence dynamically and leverage it for better exploration.
\citet{dynamic_action_repeat_extended_q} introduce a method called Dynamic Action Repetition, which extends the Q-function to include Q-values for repeated actions, allowing dynamic selection of action persistences.
However, this approach is limited to discrete action spaces and struggles when dealing with extensive action persistences.
To address these limitations, subsequent work by \citet{2017learning_to_repeat} and \citet {temporl} propose training an additional repetition policy specifically for action persistence.

Instead of providing an action persistence for each action, \pw{alternative methods have been explored.
For instance, \citet{TAAC} propose} a binary policy to decide whether to repeat \pw{the} last action \pw{or not, while}
\citet{time-discretization-invariant-SAR} propose a safe region where the agent repeats actions until leaving this region. 
The Option \citep{option_framework} framework introduces temporally extends actions, known as options.
Action persistence can be viewed as a special case called Persistent Option \citep{all_persistent_values}.
Given trajectories under an action persistence, existing methods \citep{temporally_extended_actions, q_action_persistence, all_persistent_values} can learn the value function for another action persistence or 
simultaneously for multiple action persistences.

\subsubsection*{Exploration in DRL}
Recent advancements in exploration strategies for DRL have introduced a variety of innovative methods \citep{exploration_survey}.
One prominent approach to address the dilemma between exploration and exploitation is the use of intrinsic reward, by which the agent is guided to explore. 
Curiosity-driven exploration \citep{dynamic_ae, ICM, curiosity_driven} and novelty-based rewards \citep{count_based_exploration, neural_density_model, count_with_successor_representation, rnd} fall into this category.
Curiosity-driven exploration encourages agents to visit states that maximize the error in their learned predictive models, while novelty-based rewards guide the agent to explore novel states according to some novelty measures.
Another significant advancement is the integration of uncertainty estimation into exploration strategies \citep{bayesian_dqn, bootstrapped_dqn, sunrise, mavrin2019distributional, Zhou2020NonCrossingQR}.
Uncertainty based exploration methods quantify the uncertainty, helping agents to more effectively explore areas where the model is less certain.

Moreover, recent work has also focused on improving exploration through the use of hierarchical structures \citep{hierarchical_rl} and temporal abstraction \citep{option_framework}. 
Techniques such as options and macro-actions enable agents to operate at multiple levels of abstraction, allowing for more structured and efficient exploration.
\citet{variable_time_discretization} train agents with progressively finer time discretization.
Notably, \citet{temporally_extended_epsilon} propose temporally-extended $\epsilon$-greedy which is the most relevant one to our method.
On each step, besides choosing a random action with probability $\epsilon$, a zeta distribution ($z(n) \propto n^{-\mu}$) is used to decide the action persistence of a random action.
This allows the agent to explore persistently without modifying the greedy policy.

\section{Background}\label{Background}
In this section, we introduce the notations and recall the basic formulation of the actor-critic framework in deep reinforcement learning.
\subsection{Problem Formulation}
For any set $\mathcal X$, $\Delta(\mathcal X)$ denotes the set of probability distributions over $\mathcal X$.
For a random variable $X$, $\mathbb E[X]$ (resp. $\mathbb V[X]$) denotes its expectation (resp. variance).
For any function $\phi$ and i.i.d. samples $x_1, \ldots, x_N$ of $X$, $\eExpect[\phi(X)] = \frac{1}{N} \sum_{i=1}^N \phi(x_i)$ is an empirical mean \yban{estimation} of $\mathbb E[\phi(X)]$.
Meanwhile, a Markov Decision Process (MDP) $M = (\St,\Ac, \Reward, \T, \rho_0)$ is 
composed of a set of state $\St$, a set of action $\Ac$, a reward function $\Reward: \St \times \Ac \rightarrow \mathbb{R}$, a transition function $\T:\St \times \Ac \rightarrow \Delta(\St)$, and a probability distribution over initial states $\rho_0 \in \Delta(\St)$.
In reinforcement learning, the agent is trained by interacting with the environment to learn a policy $\pi(\cdot \mid s) \in \Delta(\Ac)$ such that the expected return $ \mathbb{E}_\pi[\sum_{t=0}^\infty \gamma^tr_t\mid s_0 \sim \rho_0]$ is maximized.

\subsection{Actor-Critic Framework}
A typical \pw{off-policy} actor-critic framework in DRL parameterizes the actor $\pi_{\theta}$ and critic $Q_{\phi}$ with neural networks.
To solve the MDP, specific actor and critic losses are calculated using samples collected from a replay buffer $\mathcal{D}$ for updating the parameters.
We introduce two typical actor-critic algorithms below, which are related to the base algorithms used in the experiments.\\[1ex]
\noindent \textbf{Deep Deterministic Policy Gradient}
Incorporating a parameterized deterministic actor function $\mu_\theta(s)$, Deep Deterministic Policy Gradient (DDPG) \citep{DDPG} uses the following actor and critic loss to train the agent:
\begin{equation}
\begin{split}
    &J_\mu(\theta) = \eExpect_{s_t\sim \mathcal{D}}[-Q_\phi(s_t,a_t)|_{a_t=\mu_{\theta}(s_t)}],\\
    &J_Q(\phi) = \eExpect_{s_t,a_t,s_{t+1},r_t \sim\mathcal{D}}[(Q_\phi(s_t,a_t)-(r_t+\gamma \hat{Q}_{\bar{\phi}}(s_{t+1},a_{t+1}')))^2|_{a_{t+1}'=\mu_{\bar{\theta}}(s_{t+1})}],
\end{split} 
\end{equation}
where 
$\hat{Q}_{\bar\phi}(s_{t+1},a'_{t+1})$ is the Q-value defined from a target network, $\theta$ and $\phi$ \pw{represent} the parameters of the actor and the critic respectively, $\bar\theta$ and $\bar\phi$ \pw{represent} the parameters of the target actor and the target critic respectively, and $\mathcal D$ represents the replay buffer. The weights of a target network are the exponentially moving average of the online network's weights.
Note that noise sampled from a noise process $\mathcal{N}$ is added to the actor policy for exploration in DDPG:
\begin{equation}
    \mu'(s_t) = \mu_\theta(s_t)+\pw{\varepsilon ~ \mbox{ where } ~ \varepsilon \sim \mathcal{N}}
\end{equation}

\noindent \textbf{Soft Actor-Critic} Maximum entropy RL tackles an RL problem with an alternative objective function, which favors more random policies: $J = \eExpect_\pi[\sum_{t=0}^\infty \gamma^t r_t+\alpha \mathcal{H}(\pi(\cdot \mid s_t))]$, where $\gamma$ is the discount factor, $\alpha$ is a trainable coefficient of the entropy term and $\mathcal{H}(\pi(\cdot \mid s_t))$ is the entropy of action distribution $\pi(\cdot \mid s_t)$. 
The Soft Actor-Critic (SAC) algorithm \citep{SAC} optimizes it by training the actor $\pi_\theta$ and critic $Q_\phi$ with the following losses:
\begin{equation}
\begin{split}
    &J_\pi(\theta) = \eExpect_{s_t\sim \mathcal{D},a \sim \pi_\theta(s_t)}[\alpha\log \pi_\theta(a \mid s_t)-Q_\phi(s_t,a)],\\
    &J_Q(\phi) = \eExpect_{s_t,a_t,s_{t+1},r_t \sim \mathcal{D}}[(Q_\phi(s_t,a_t)-\\
    &\qquad\qquad (r_t+\gamma \hat{Q}_{\bar{\phi}}(s_{t+1},a_{t+1}')-\alpha\log{\pi_{\theta}(a_{t+1}'|s_{t+1})})
    )^2|_{a_{t+1}'\sim\pi_\theta(s_{t+1})}].
\end{split} 
\end{equation}
The same notations used in DDPG above, such as the parameters for the actor and critic are used here as well.

\section{Effect of Action Persistence}
In this section, we analyze the effect of both large and small persistence in DRL algorithms. Intuitively, larger action persistences mean fewer decisions required in a trajectory.
Also, larger action persistences contribute to broader exploration by sticking to temporally persistent action sequences.
Accordingly, smaller action persistences correspond to fine-grained policies which usually yield superior optimal performance.

\begin{figure}[tb]
\centering
\begin{subfigure}[b]{0.32\textwidth}
\includegraphics[width=\linewidth]{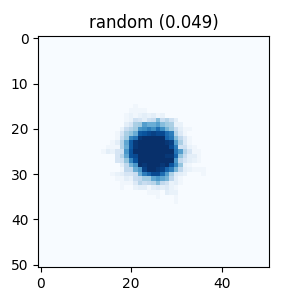}
\end{subfigure}
\begin{subfigure}[b]{0.32\textwidth}
\includegraphics[width=\linewidth]{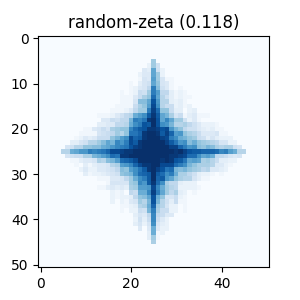}
\end{subfigure}
\begin{subfigure}[b]{0.32\textwidth}
\includegraphics[width=\linewidth]{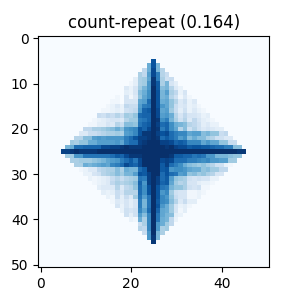}
\end{subfigure}
\begin{subfigure}[b]{0.32\textwidth}
\includegraphics[width=\linewidth]{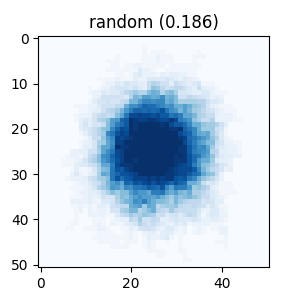}
\end{subfigure}
\begin{subfigure}[b]{0.32\textwidth}
\includegraphics[width=\linewidth]{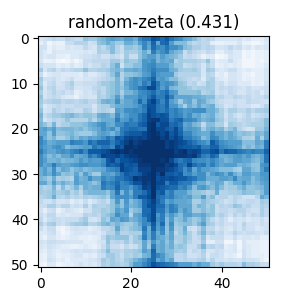}
\end{subfigure}
\begin{subfigure}[b]{0.32\textwidth}
\includegraphics[width=\linewidth]{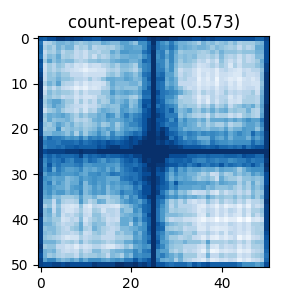}
\end{subfigure}
\caption{\textbf{State coverage of executing different policies in the mini grid.}
The deeper color in the grid corresponds to larger probability to visit the state.
The episode length and total time step in one run are set as (20, 1000) for the results in the first row and (100, 3000) for the results in the second row. 
The percentage of state coverage averaged over 30 runs is indicated in the title of each plot.}
\label{fig:pure exploration and coverage of state space}
\end{figure}

\subsection{Large Action Persistence}
As observed in previous work, using large action persistence accelerate the exploration of the state space.
We provide some theoretical formalization to explain this observation.
When using a large action persistence, the \pw{number of} action decisions made in a finite horizon \pw{is} smaller, leading to fewer possible action sequences.
For \pw{an} MDP with a finite horizon $H$, with a larger action persistence, the size of the set of all possible action sequences of length $H$ is smaller:
\begin{equation}
    \|\Ac_{\ap_1}\|<\|\Ac_{\ap_2}\|, \text{ if } \ap_1>\ap_2.
\end{equation}
where $\Ac_{\ap}$ is the set of all possible action sequences of length $H$ with an action persistence $\ap \in \aps$: 
\begin{equation}
\Ac_{\ap}=\{(\underbrace{a_0,...,a_0}_{\ap},...,\underbrace{a_{h-1},...,a_{h-1}}_{\ap},\underbrace{a_h,...,a_h}_{H-\ap*(h-1)}) \mid a_i\in \Ac, \forall i = 0,...,h \}
\end{equation}
with $h=\left\lceil \frac{H}{\ap} \right\rceil$ is the number of decisions made in an action sequences from $\Ac_\ap$.

Given two sets $\|\Ac_{\ap_1}\|$ and $\|\Ac_{\ap_2}\|$, which are both sufficient to cover the whole state space, exploration by trying random actions will be more efficient by considering the larger action persistence.
However, using a large action persistence may prevent to cover the whole state space. 
In such case, a stochastic action persistence may circumvent this issue, while still preserving some benefits from large action persistence \citep{temporally_extended_epsilon}.
In our work, we further demonstrate that sampling the action persistence using a distribution that depends on both states and past exploration can improve further the exploration efficiency.

%
To further illustrate that incorporating repeated actions promotes temporally persistent exploration, we evaluate different behavior policies with or without repeated actions in a $51 \times 51$ mini grid environment.
In this setting, the state space $\St \pw{=} \{(i, j) \mid 0 \leq i \leq 50, 0\leq j \leq 50\}$ contains all grid coordinates, and the agent can select actions from the action space $\Ac \pw{=} \{up, down, left, right\}$ to move one step in the corresponding direction until reaching the boundaries.
When a boundary is reached, any action attempting to move the agent outside the grid does not change its state, while other actions proceed normally.
The agent is initialized at the center of the grid and is also reset to the center at the end of each episode.
The policies tested in this mini-grid are as follows:
\begin{enumerate}
    \item \textbf{Random\pw{:}} Actions are uniformly sampled from the action space.
    \item \textbf{Random-zeta\pw{:}} 
    As proposed in temporally-extended $\epsilon$-greedy \citep{temporally_extended_epsilon}, a fixed zeta-distribution ($z(n) \propto n^{-\mu}$) is used for deciding action persistence of a random action. 
    We adopt the hyperparameter $\mu=2$ from $\epsilon z$-greedy \citep{temporally_extended_epsilon}.
    \item \textbf{Count-repeat\pw{:}} 
    Our method is instantiated for this environment.
    To specify, our method counts the visiting times of each state and decides the probability of repeating the last action.
    Higher probabilities are assigned to novel states.
    A detailed description of our method is presented in \Cref{Methodology}.
\end{enumerate}
\Cref{fig:pure exploration and coverage of state space} visualizes the state coverage achieved by rolling out trajectories, with results averaged over 30 runs.
Each run maintains the same episode length and total time steps.
This visualization clearly demonstrates that incorporating repeated actions into the exploratory policy results in broader exploration.
Interestingly, our method can achieve greater state coverage within a fixed time duration.

\subsection{Small Action Persistence}
Repeating actions encourages exploration, while smaller action persistences hold the potential for achieving better optimal performance.
For a finite horizon MDP, a policy with an action persistence $\ap_1$ can always recover a policy with an action persistence $\ap_2$, provided that $\ap_1$ is a factor of $\ap_2$.
Therefore, the set $\Pi_{\ap_1}$ of policies with an action persistence $\ap_1$ is larger than or equal to the set $\Pi_{\ap_2}$ of policies with an action persistence $\ap_2$.
Correspondingly, the expected return of the optimal policy in $\Pi_{\ap_1}$ is greater than or equal to the expected return of the optimal policy in $\Pi_{\ap_2}$:
\begin{equation}
\max_{\pi \in \Pi_{\ap_1}} \Expect_\pi \sum_{t=0}^H \gamma^tr_t\mid s_0 \sim \rho_0
\geq \max_{\pi \in \Pi_{\ap_2}} \Expect_\pi \sum_{t=0}^H \gamma^tr_t\mid s_0 \sim \rho_0.
\end{equation}
where $\ap_1$ is a factor of $\ap_2$.
There is a special case of $\ap_1=1$ where the corresponding optimal return is greater than or equal to the return of the optimal policy with any action persistence. 



\begin{figure}[t!]
    \centering
    \includegraphics[width=1.0\linewidth]{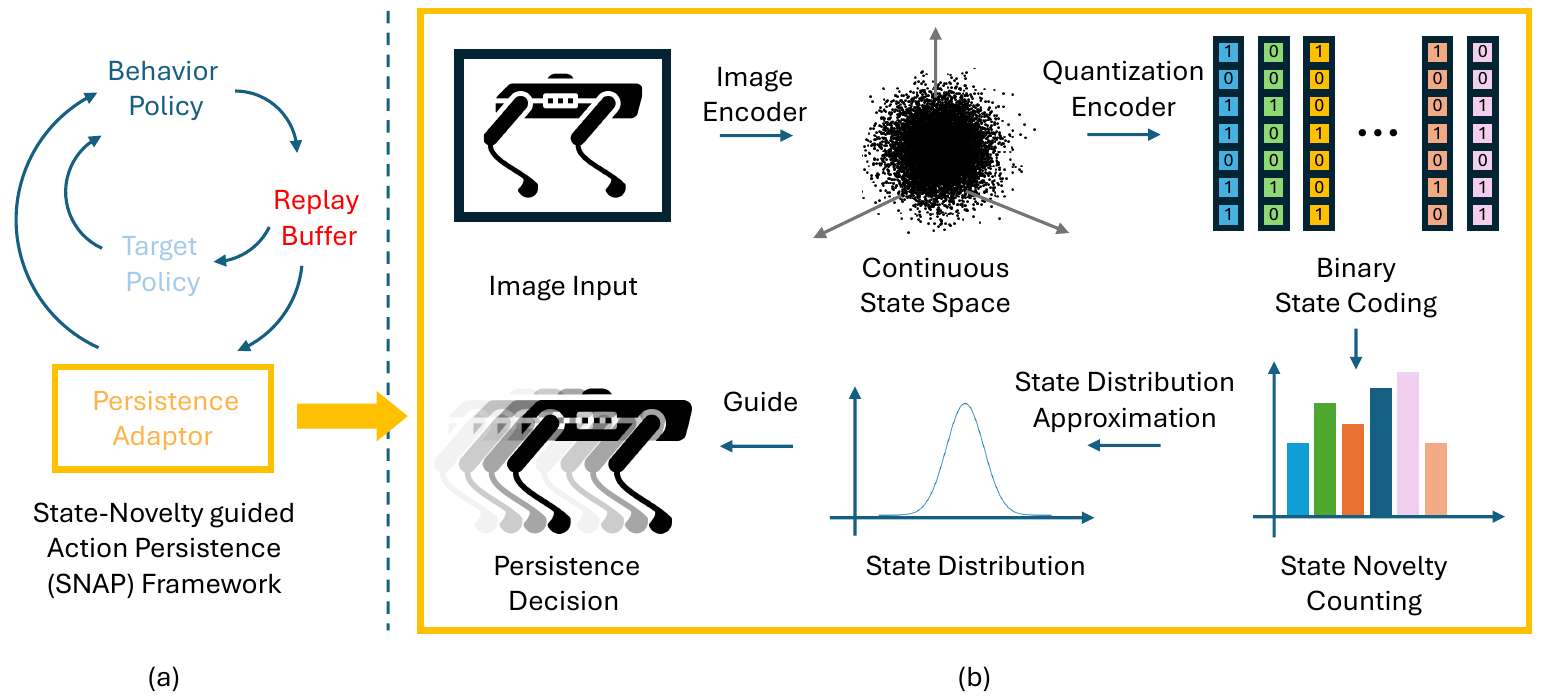}
    \caption{
    \textbf{Overview.}
    (a) An additional action persistence adaptor, guided by state-novelty, is integrated to an off-policy DRL framework.
    (b) Measure novelty directly from image input as the state is challenging.
    To address this, an image encoder is used to convert the images into feature vectors, which are then mapped to binary codes by a quantization encoder.
    By counting with the binary codes, we approximate the state distribution of the training data used for the actor or actor-critic.
    It guides the action persistence in the behavior policy.} 
    \label{fig:teaser2}
    \vspace{-3ex}
\end{figure}

In summary, we have an intuitive understanding of how incorporating repeated actions into the policy influences exploration and the optimal return.
Repeating actions leads to temporally persistent exploration while a small action persistence usually yields a better optimal return.
Next, we introduce a method of dynamically choosing action persistences in \Cref{Methodology}, followed by experimental results evaluating different ways of exploiting action persistences in \Cref{Experiemenal Results}.

\section{Methodology}\label{Methodology}

\begin{algorithm}
\caption{SNAP algorithm, where changes with respect to standard off-policy actor-critic algorithms are colored orange.}\label{algo1}
\label{alg:1}
\begin{algorithmic}[1]
\Require total training steps $\mathcal{T}$, mini-batch size $N$, \textcolor{orange}{coefficient $\alpha$} in \Cref{eq:probability of repeating}
\State Initialize encoder $g_\psi$, critic $Q_\phi(s, a)$, actor $\pi_\theta(s)$, empty replay buffer $\mathcal{D}$, \textcolor{orange}{projection matrix $A$} and \textcolor{orange}{hash table $\tilde N(\cdot)$}
\State Start with initial state $s_0$.
\For{$t=0 \dots T$}
    \If {at initial state $s_0$} 
    \State Interact with the environment using action from target policy $a_t \sim \pi(\cdot\mid s_0)$.
    \Else
    \State \textcolor{orange}{Calculate the repeat probability using \Cref{eq:probability of repeating}}
    \If {repeat}
    \State \textcolor{orange}{Interact with the environment using last action $a_t = a'$.}
    \Else     
    \State Interact with the environment using action from target policy $a_t \sim \pi(\cdot\mid s_t)$.
    \EndIf
    \EndIf
    \State \textcolor{orange}{Save the action $a'=a_t$} and save transition $(s_t,a_t,r_t,s_{t+1})$ in replay buffer $\mathcal D$.
    \State // Actor-critic update and \textcolor{orange}{hash table update}
    \State Sample mini-batch $\{(s_i, a_i, r_i, s'_i) \mid i=1, \ldots, N \}$ from $\mathcal{D}$.
    \State Update $\psi$ (encoder parameters), $\phi$ (critic parameters) and $\theta$ (actor parameters)
    \State \textcolor{orange}{Update hash table $\tilde{N}(\cdot)$ using the mini-batch} with \Cref{eq:hash function}
    \If {episode ends} 
    \State Reset $s_{t+1}$ to an initial state
    \EndIf
\EndFor
\end{algorithmic}
\end{algorithm}

According to our analysis, incorporating large action persistences is an intuitive and effective way to achieve temporally persistent exploration. Meanwhile, smaller action persistences usually result in preferable fine-grained final optimal policies.
To effectively balance the exploration and exploitation, we propose State-Novelty guided Action Persistence (SNAP). SNAP allows us to dynamically bridge the gap between large persistence strategies and small persistence strategies.
In this section, we present (i) SNAP, which is an off-policy framework of adapting dynamic action persistences in the behavior policy,
(ii) a state-novelty guided persistence adaptor to decide action persistence, (iii) and an instantiation of our method in an actor-critic algorithm.

\subsection{SNAP Framework}
This framework exploits the strong exploration capability of policies with large action persistences at the initial training stages and achieves superior performance of the fine-grained policies as training progresses.
To achieve this, a state-novelty guided adaptor to decide action persistence is introduced in the behavior policy of an off-policy framework.
By decoupling the behavior policy from the target policy, the off-policy framework allows for diverse exploration strategies without compromising the stability of the training process.
Consequently, it enables leveraging different action persistences for varying levels of exploration.
Meanwhile, it does not require to train additional value functions or policies specifically for action persistence decisions anymore.

Ideally, action persistence should gradually transition from large to small.
At the beginning of the training, large action persistences facilitate temporally persistent exploration, which leads to a larger coverage of the state space.
As training progresses, action persistence is expected to gradually decrease for obtaining a fine-grained policy.
Ultimately, the action persistence in the behavior policy should closely align with that of the target policy.
Therefore, we choose to use state-novelty, a measure to quantitatively evaluate how familiar the agent is with the current state.
It is ideal for guiding action persistence since it naturally decreases as the agent explores. 

\subsection{Persistence Adaptor}
In this part, we introduce the persistence adaptor module which dynamically makes the action persistence decision in behavior policy.
Our method employs a count-based measure \citep{count_based_exploration} for state-novelty.
An illustration of the persistence adaptor is shown in \Cref{fig:teaser2} (b).
In this module, we have the following components.
Firstly, to mitigate the dimensional explosion caused by using image inputs as states, an image encoder is applied to convert the images into feature vectors.
In an image-based DRL architecture, there is an image encoder, trained by the critic loss, to map images to meaningful representations.
It is reasonable to leverage this representation by sharing the encoder with the adaptor.
Next, we use a hash function \citep{simhash} as a quantization encoder to map the feature space to binary code space.
The hashed representation has three main advantages: (i) The hashed space allows us to measure the state-novelty by counting; 
(ii) Close feature vectors are usually mapped to the same binary code;
and (iii) The hash function is computationally efficient. The hash function $\phi(s)$ used can be formally written as:
\begin{equation}
    \phi(s) = \textbf{\text{sgn}}(A \cdot g(s)) \in \{1,-1\}^K,
\label{eq:hash function}
\end{equation}
where g is the encoder $g:\mathcal{S} \rightarrow \mathbb{R}^D$, $A$ is a $K \times D$ matrix with elements drawn from a standard Gaussian distribution $\mathcal{N}(0,1)$ and \textbf{sgn} corresponds to a sign function.
To adaptively choose action persistence, the adaptor decides the probability of repeating last action for each state.
The probability of repeating the previous action $a_{t-1}$ for state $s_t$ in the behavior policy $\pi'$ is given by
\begin{equation}
    P\big(\pi'(s_t)=a_{t-1}\big) = \frac{\alpha}{\pw{\max}(1, \sqrt{\tilde{N}(s_t)})},
\label{eq:probability of repeating}
\end{equation}
where $\alpha$ is a hyperparameter controlling exploration and $\tilde{N}(s_t)$ is a pseudo-count, counted with the binary codes of the state $s_t$.
The proposed formula for calculating the probabilities of action repetition allows for dynamic adjustment of action persistence, adapting progressively throughout the training process.

\subsection{Algorithm}
\Cref{alg:1} integrates the proposed SNAP into an off-policy actor-critic framework.
The difference between the SNAP algorithm and the traditional actor-critic algorithm is highlighted in orange, which contains two key modifications:
\begin{enumerate}
    \item It adopts the persistence adaptor to decide the action persistence in behavior policy, while interacting with the environments.
    \item It dynamically updates the state-novelty measure by using samples from the replay buffer.
\end{enumerate}
Note that the hash table is not updated immediately upon encountering new states.
Instead, we update the hash table using the same mini-batch used for training the actor and critic.
Consequently, this count-based method approximates the state distribution of the training data for the actor and critic.
This provides an effective measure of the familiarity of the agent with different states.
Moreover, this instantiation clearly demonstrates that our method can be seamlessly integrated into various off-policy algorithms in DRL.

\section{Experimental Results}\label{Experiemenal Results}
To validate our proposition and show the effectiveness of our proposed algorithm, we conduct the following experiments: 
\begin{itemize}
    \item We experimentally demonstrate the effect of using different action persistences;
    \item We benchmark different methods of exploiting repeated actions;
    \item We evaluate the performance of combining the proposed SNAP with different basic exploration strategies: \textbf{Epsilon-greedy, Entropy-regularization} and \textbf{Noisy net}.
    \item 
    {We conduct an ablation study to assess the design choice of the probability scheduler and the measure of state novelty.}
\end{itemize}

\subsubsection*{Experimental Set Up}
\begin{figure}[tb]
\centering
\begin{subfigure}[b]{0.32\textwidth}
\includegraphics[width=\linewidth]{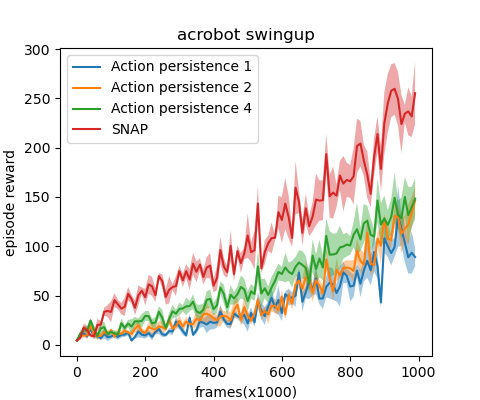}
\end{subfigure}
\begin{subfigure}[b]{0.32\textwidth}
\includegraphics[width=\linewidth]{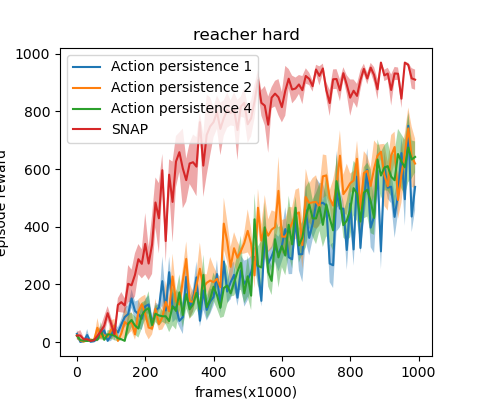}
\end{subfigure}
\begin{subfigure}[b]{0.32\textwidth}
\includegraphics[width=\linewidth]{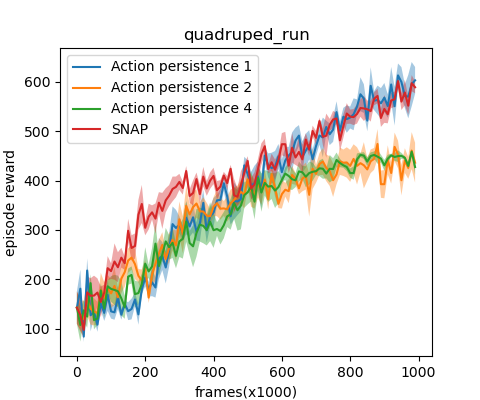}
\end{subfigure}
\begin{subfigure}[b]{0.32\textwidth}
\includegraphics[width=\linewidth]{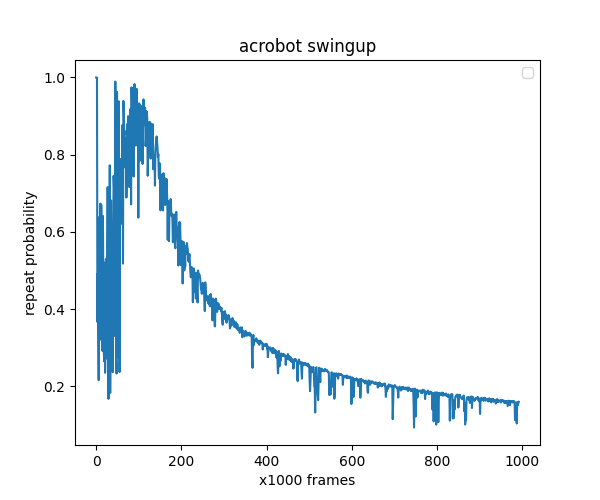}
\end{subfigure}
\begin{subfigure}[b]{0.32\textwidth}
\includegraphics[width=\linewidth]{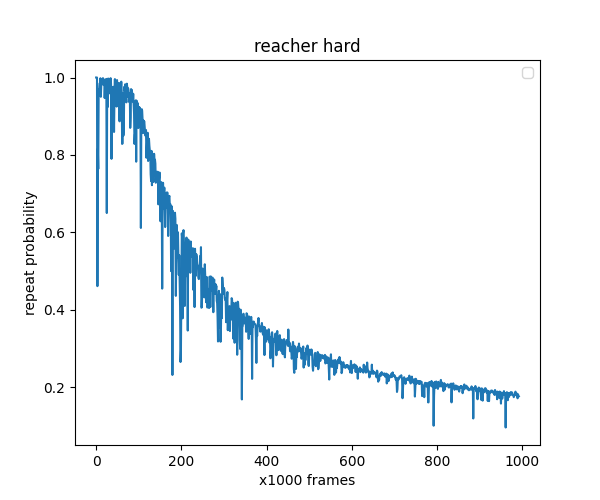}
\end{subfigure}
\begin{subfigure}[b]{0.32\textwidth}
\includegraphics[width=\linewidth]{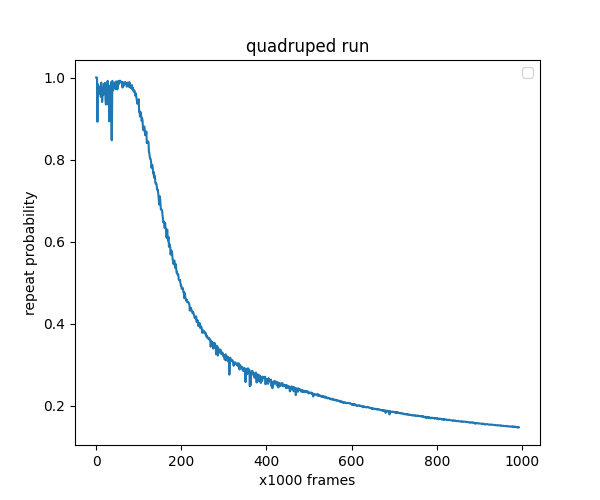}
\end{subfigure}
\caption{\textbf{DrQv2 with different action persistences.} The results of comparing SNAP with using a fixed action persistence are shown in the first row.
The sample efficiency of our method can not be achieved by simply tuning the action persistence as a hyperparameter.
The probabilities averaged over 1000 frames are shown in the second row.
Initially, a high probability of repeating actions enables temporally persistent exploration.
As training progresses, this probability decreases, ensuring a fine-grained policy is employed in the later stages.}
\label{fig:different repeat}
\end{figure}

We evaluate the sample efficiency of different methods on environments from \textbf{DeepMind Control Suite} (DMControl) \citep{dm_control}.
It is a commonly used benchmark for evaluating DRL algorithms.
DMControl contains many robotics tasks with different dimensions of state space and action space.
We focus on image-based environments and use state-of-the-art DRL algorithm called DrQv2 as our basic algorithm.
By leveraging data augmentation, DrQv2 is known for the promising sample efficiency in image-based control tasks.
Across different environments, all hyperparameters are listed in 
\Cref{appendix:hyperparameters} 
such as learning rates and batch size for the actor and critic. 
Without specification, most of the hyperparameters are kept the same as those used in the original DrQv2.
Note that the default action persistence is set to 2 in DrQv2.
All experiments are performed with 5 different random seeds and the agent is evaluated every 10k environment steps, whose performance is measured by cumulative rewards averaged over 10 evaluation episodes.

\subsection{Different Action Persistences}
\begin{figure}[tb]
\centering
\begin{subfigure}[b]{0.32\textwidth}
\includegraphics[width=\linewidth]{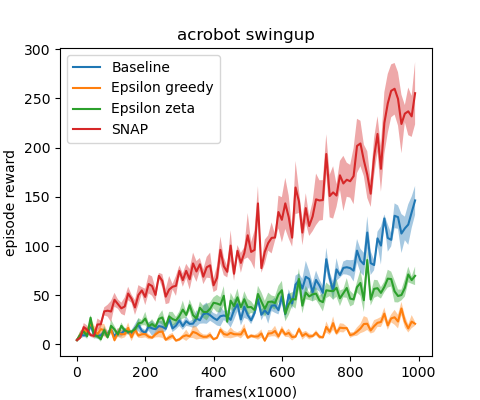}
\end{subfigure}
\begin{subfigure}[b]{0.32\textwidth}
\includegraphics[width=\linewidth]{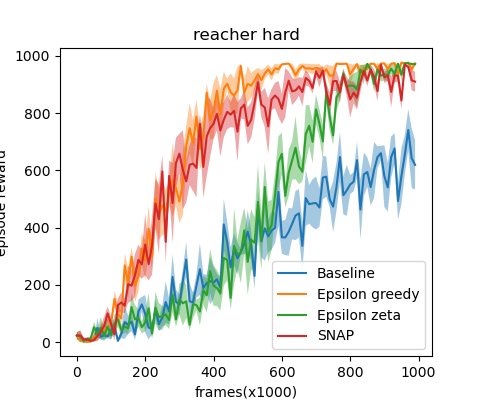}
\end{subfigure}
\begin{subfigure}[b]{0.32\textwidth}
\includegraphics[width=\linewidth]{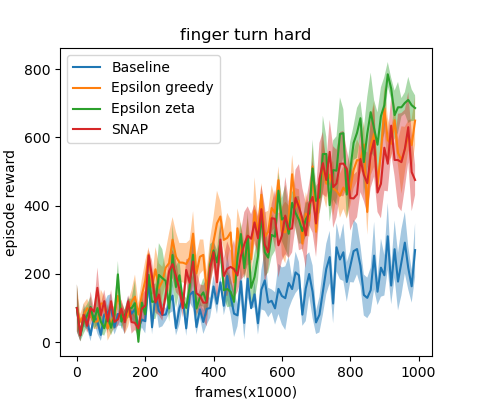}
\end{subfigure}
\begin{subfigure}[b]{0.32\textwidth}
\includegraphics[width=\linewidth]{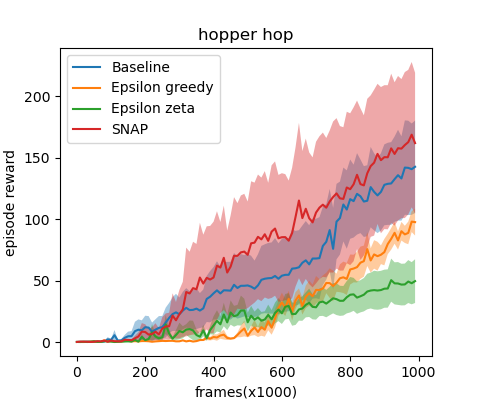}
\end{subfigure}
\begin{subfigure}[b]{0.32\textwidth}
\includegraphics[width=\linewidth]{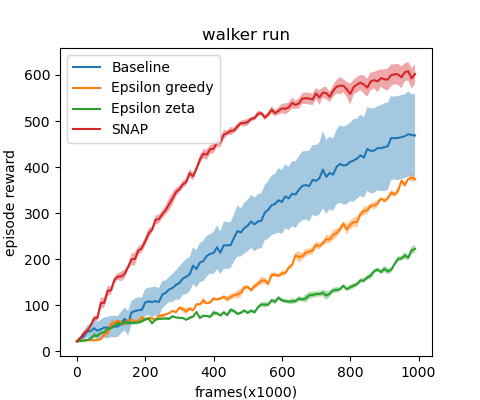}
\end{subfigure}
\begin{subfigure}[b]{0.32\textwidth}
\includegraphics[width=\linewidth]{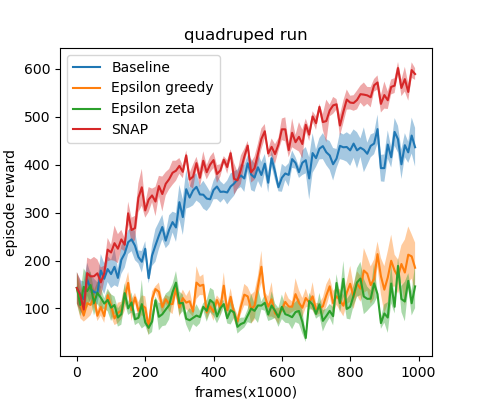}
\end{subfigure}
\begin{subfigure}[b]{0.32\textwidth}
\includegraphics[width=\linewidth]{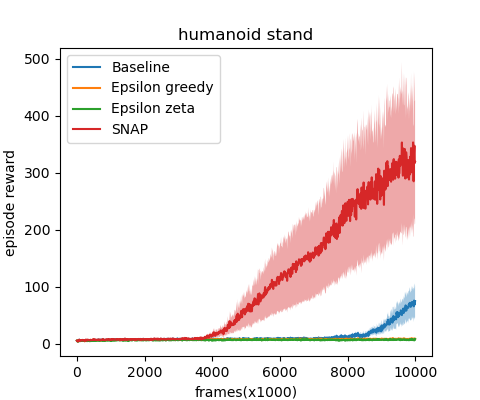}
\end{subfigure}
\begin{subfigure}[b]{0.32\textwidth}
\includegraphics[width=\linewidth]{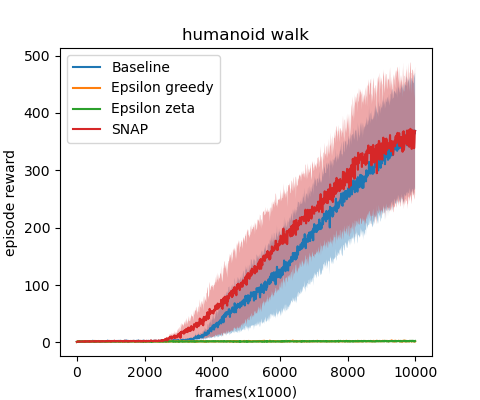}
\end{subfigure}
\begin{subfigure}[b]{0.32\textwidth}
\includegraphics[width=\linewidth]{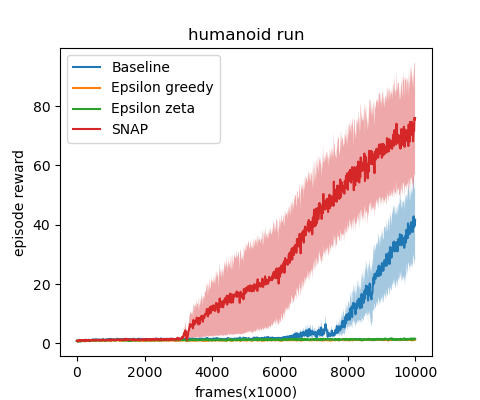}
\end{subfigure}
\caption{\textbf{Performance of incorporating repeated actions.} 
Two ways of incorporating temporally persistent exploration are compared with the baseline.
Epsilon-zeta simply uses a fixed zeta distribution to decide the number of time steps for repeating the random actions.
Our method dynamically determines repeating actions based on the state-novelty.}
\label{fig:all performance}
\end{figure}
We start with experiments of using different action persistences $\ap={1,2,4}$ in Reacher-hard, Acrobot-swingup and Quadruped-run, as shown in \Cref{fig:different repeat}.
{The additional results of training the agents until convergence are shown in \Cref{appendix:additional_exps}.}
The dimensions of state space and action space of these three environments are respectively (4,1), (4,2) and (56, 12).
Quadruped-run is a hard locomotion task due to the high dimensions and rich contact.

From the results, it is observed that the choice of action persistence influences the sample efficiency of the original algorithm (DrQv2).
There is not a fixed action persistence which can be optimal across all the environments.
Compared to using a fixed action persistence, state-dependent action persistences in our method lead to a significant gain in sample efficiency.
This can not be achieved by tuning the action persistence as a hyperparameter specifically for each environment.
To concretely present how our method adjusts the action persistence during the training, the probabilities of repeating actions for all steps are saved.
Based on our count-based method, this smooth scheduler for the repeat probabilities enables both the efficiency in exploration and the enhancement in performance.

\subsection{Exploiting Repeated Actions}

\begin{figure}[tb]
\centering
\begin{subfigure}[b]{0.95\textwidth}
\includegraphics[width=\linewidth]{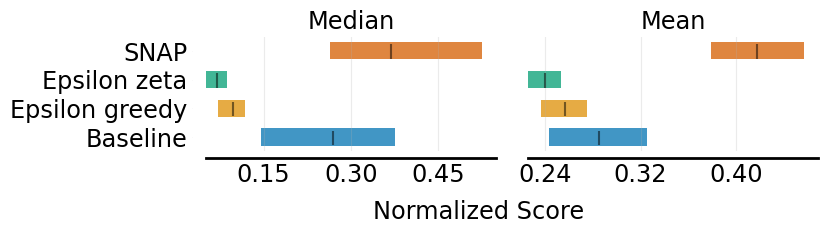}
\end{subfigure}
\begin{subfigure}[b]{0.95\textwidth}
\includegraphics[width=\linewidth]{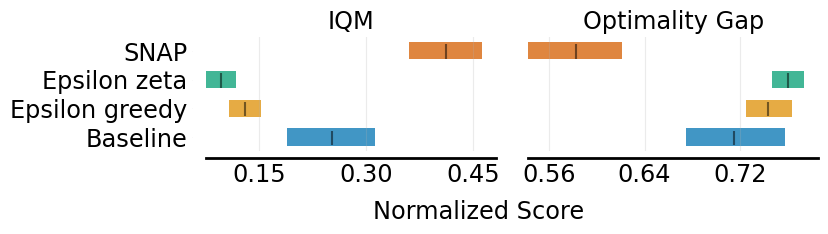}
\end{subfigure}
\caption{\textbf{Aggregated Performance.} 
The episode rewards are normalized by 1000 in all environments.
Several statistics are calculated using the normalized scores for evaluating the aggregated performance.
Better performance is indicated by higher median, IQM, and mean values, and a lower Optimality Gap.}
\label{fig:aggregated performance}
\end{figure}

The experiments above provide some insights of how different action persistences influence the exploration and the sample efficiency.
They also partially demonstrate the benefits of incorporating temporally persistent exploration into the policy.
Here, we further compare ours with another method of exploiting repeated actions, called \textbf{temporally-extended $\epsilon$-greedy}.
The comprehensive evaluations across tasks with varying complexities are shown in \Cref{fig:all performance}.
Following the recommendations from \citet{aggregated_performance}, the aggregated performance is also plotted in \Cref{fig:aggregated performance}.
As expected, employing a fixed distribution to decide the action persistence proves ineffective in complex tasks, especially in those contact-rich tasks.  
In contrast, our method of dynamically choosing action persistences for each state can facilitate broader exploration during the initial stages and achieve higher returns at the end.

\subsection{Improving Exploration Strategies with SNAP}
\begin{figure}[tb]
\centering
\begin{subfigure}[b]{0.32\textwidth}
\includegraphics[width=\linewidth]{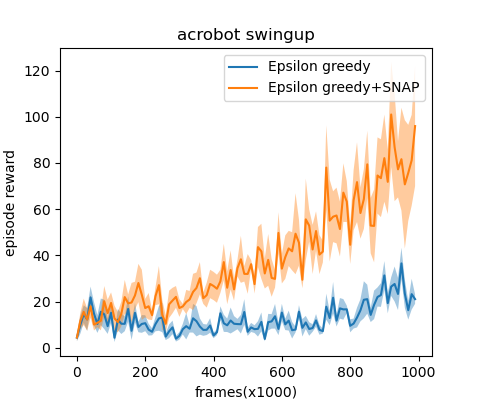}
\end{subfigure}
\begin{subfigure}[b]{0.32\textwidth}
\includegraphics[width=\linewidth]{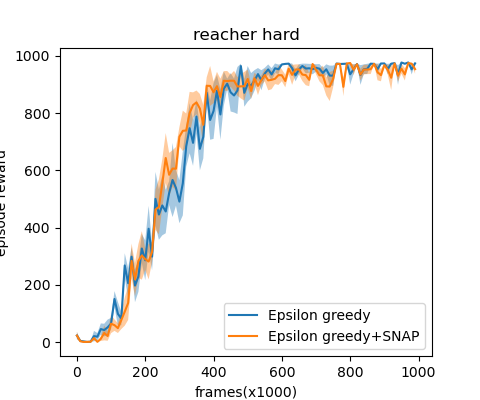}
\end{subfigure}
\begin{subfigure}[b]{0.32\textwidth}
\includegraphics[width=\linewidth]{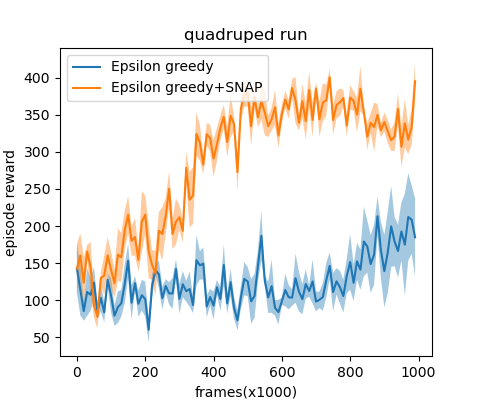}
\end{subfigure}
\begin{subfigure}[b]{0.32\textwidth}
\includegraphics[width=\linewidth]{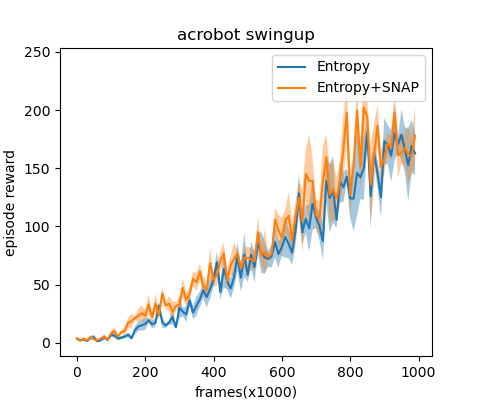}
\end{subfigure}
\begin{subfigure}[b]{0.32\textwidth}
\includegraphics[width=\linewidth]{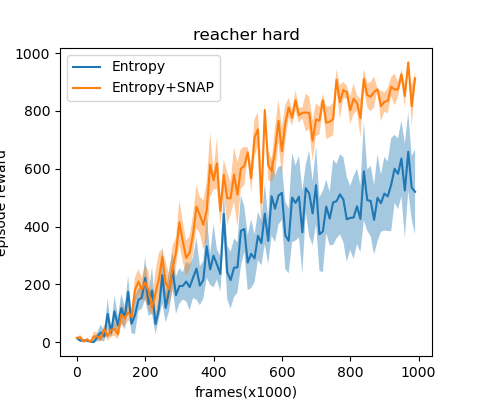}
\end{subfigure}
\begin{subfigure}[b]{0.32\textwidth}
\includegraphics[width=\linewidth]{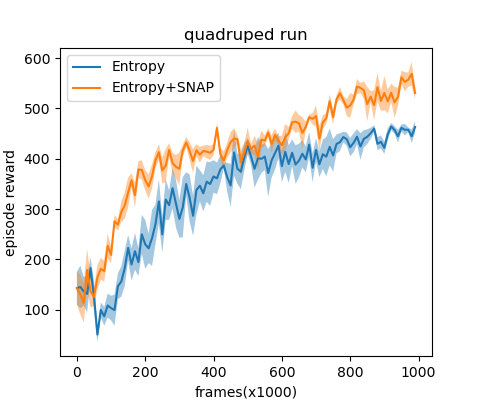}
\end{subfigure}
\begin{subfigure}[b]{0.32\textwidth}
\includegraphics[width=\linewidth]{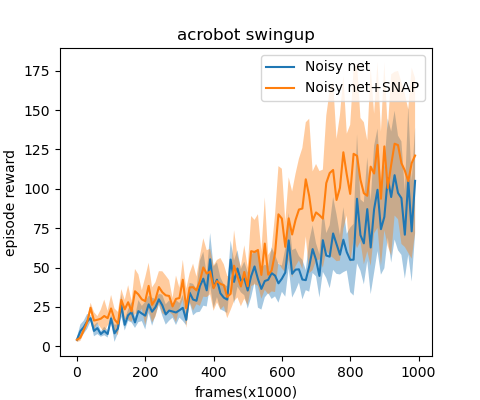}
\end{subfigure}
\begin{subfigure}[b]{0.32\textwidth}
\includegraphics[width=\linewidth]{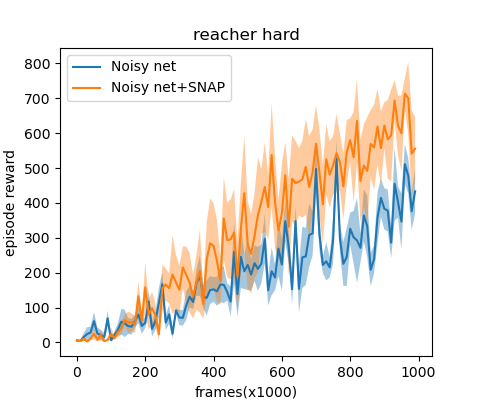}
\end{subfigure}
\begin{subfigure}[b]{0.32\textwidth}
\includegraphics[width=\linewidth]{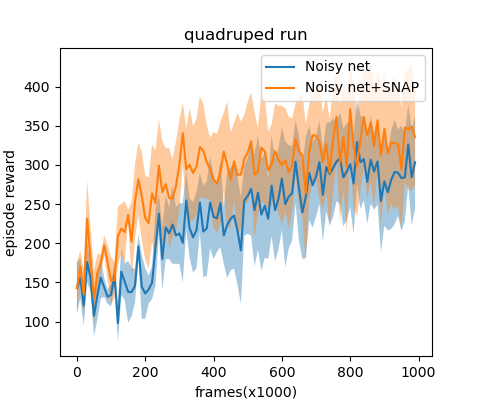}
\end{subfigure}
\caption{\textbf{Combining SNAP with several basic exploration strategies.}
The evaluated sum of rewards of combining our method with other basic exploration strategies are shown in the plots.
Our method consistently improve them in sample efficiency}
\label{fig:different exploration}
\end{figure}
As mentioned in \Cref{Background}, the base algorithm of DrQv2 is DDPG, which adds Gaussian noise to the actions from the deterministic policy to explore.
Actually, other basic exploration strategies can also be temporally extended by integrating our method with them.
We test combining our method with 
\begin{enumerate}
    \item \textbf{Epsilon greedy:} There is a probability of $\epsilon$ for using random actions.
    \item \textbf{Entropy regularization:} Replacing DDPG with SAC, which uses a stochastic policy and includes an entropy term to regularize the policy for encouraging exploration.
    \item \textbf{Noisy net:} Stochasticity is directly embedded into the neural network \citep{noisy_net} for exploration.
\end{enumerate}
These methods are widely used due to their simplicity and effectiveness without requiring domain-specific knowledge.
As shown in \Cref{fig:different exploration}, those basic exploration strategies can be temporally extended by our method, and our method consistently improves them in sample efficiency.

\begin{figure}[tb]
\centering
\begin{subfigure}[b]{0.32\textwidth}
\includegraphics[width=\linewidth]{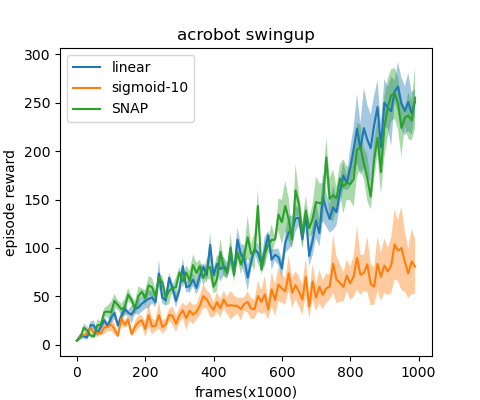}
\end{subfigure}
\begin{subfigure}[b]{0.32\textwidth}
\includegraphics[width=\linewidth]{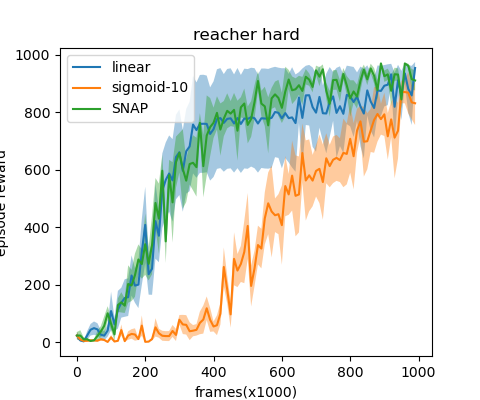}
\end{subfigure}
\begin{subfigure}[b]{0.32\textwidth}
\includegraphics[width=\linewidth]{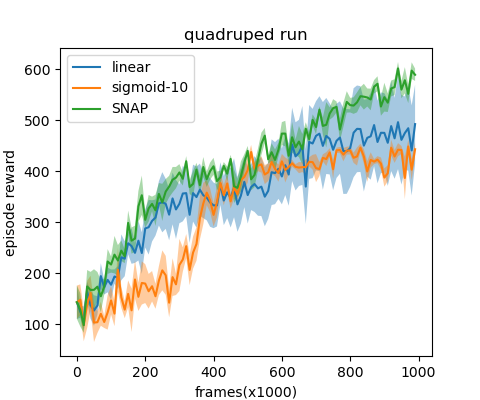}
\end{subfigure}
\begin{subfigure}[b]{0.32\textwidth}
\includegraphics[width=\linewidth]{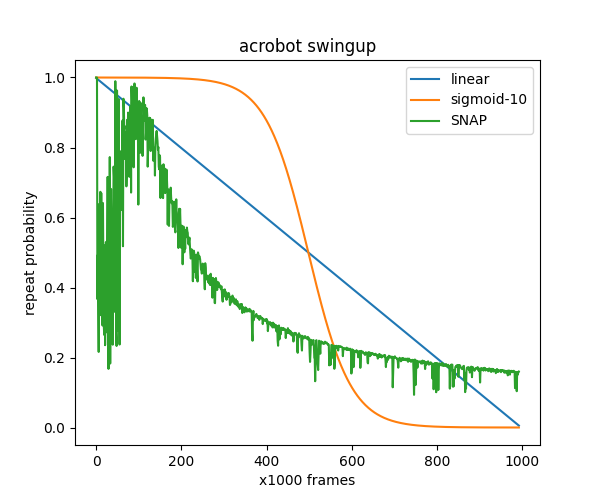}
\end{subfigure}
\begin{subfigure}[b]{0.32\textwidth}
\includegraphics[width=\linewidth]{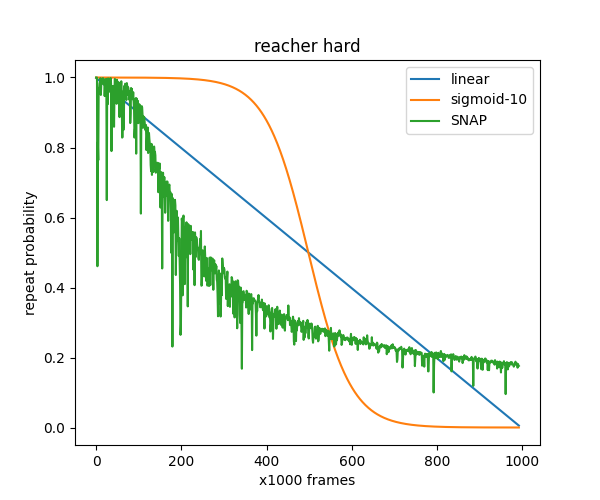}
\end{subfigure}
\begin{subfigure}[b]{0.32\textwidth}
\includegraphics[width=\linewidth]{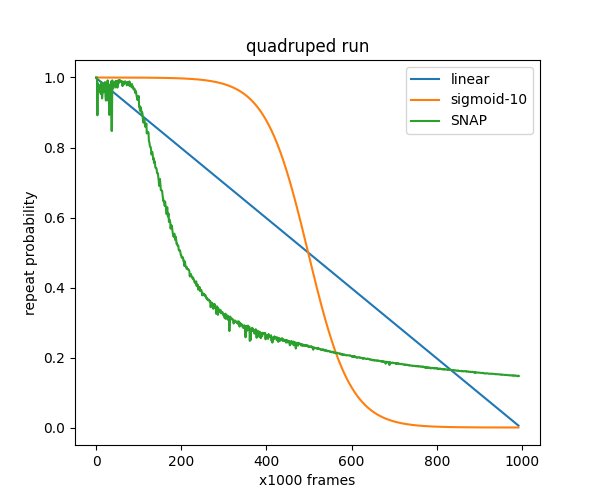}
\end{subfigure}
\caption{
{
\textbf{DrQv2 with different probability schedulers.} 
The figures show the results of using different probability scheduler: 1) linear scheduler 2) sigmoid scheduler 3) probabilities calculated by a state novelty measure.
}}
\label{fig:scheduler}
\end{figure}

\begin{figure}[tb]
\centering
\begin{subfigure}[b]{0.32\textwidth}
\includegraphics[width=\linewidth]{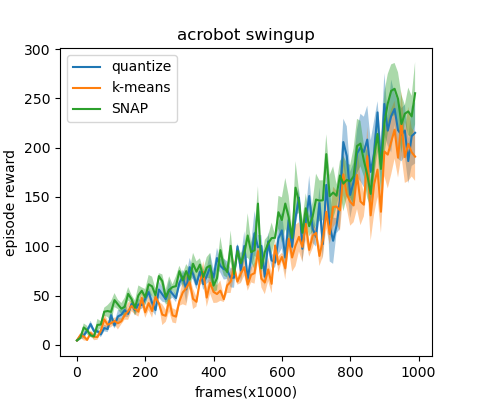}
\end{subfigure}
\begin{subfigure}[b]{0.32\textwidth}
\includegraphics[width=\linewidth]{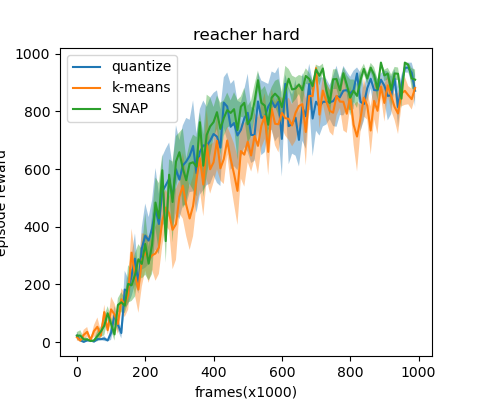}
\end{subfigure}
\begin{subfigure}[b]{0.32\textwidth}
\includegraphics[width=\linewidth]{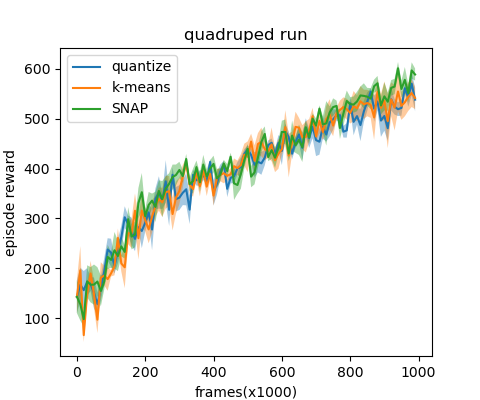}
\end{subfigure}
\caption{
{\textbf{DrQv2 with different measures of state novelty.}
To measure the state novelty, the visited states are counted by using quantized state vectors, k-means clustered states and simhash.}}
\label{fig:different_count}
\end{figure}

{
\subsection{Abalation study}
In this section, we analyze the design choice of leveraging state novelty to guide the action persistence and compare different measures of state novelty.
Firstly, a linear scheduler and a sigmoid scheduler to decide the probability of repeating actions are tested.
Both the evaluation curves and corresponding probabilities along the training are shown in \Cref{fig:scheduler}.
Our method outperforms the baselines of using a simple scheduler for deciding the probability.
Notably, our method has a smaller variance as well.
Additionally, different measures of the state novelty are compared:
1) count with quantized states 2) count with clustered states using k-means.
For quantized states, each dimension in the state vectors is first discretized and then used for counting.
As for k-means clustered states, the center of each cluster is first initialized by batches of states from the replay buffer and then continues to be updated during the training.
The results shown in \Cref{fig:different_count} indicate the robustness of our method with respect to the choice of measure for state novelty.
}

\section{Conclusion}\label{Conclusion}
In this paper, we introduce a novel method for dynamically deciding action persistence in DRL, guided by a state-novelty measure.
By incorporating temporal persistent exploration into the behavior policy of off-policy DRL algorithms, our approach balances exploration and exploitation without incurring significant computational overhead.
Experimental results on DMControl tasks demonstrated significant improvements in sample efficiency over baseline methods.
Meanwhile, our method can be integrated into other basic exploration strategies for incorporating temporal persistence.
Future work could explore the application of our method to other DRL algorithms and more environments.
Additionally, investigating alternative measures of state-novelty and their impact on action persistence strategies could provide further insights and enhance the effectiveness of our approach.

\backmatter


\begin{declaration}
\section*{Declarations}


\textbf{Funding} 
\noindent
This work has been supported by the program of National Natural Science Foundation of China (No. 62176154) and by Shanghai Magnolia Funding Pujiang Program (No. 23PJ1404400).

\noindent
\textbf{Competing Interests}
The authors declare no competing interests.

\noindent
\textbf{Ethics approval}
Not applicable

\noindent
\textbf{Consent to participate}
All authors give their consent to participate.

\noindent
\textbf{Consent for publication}
All authors give their consent for publication.

\noindent
\textbf{Code availability}
The code of the paper is available: \url{https://github.com/Jianshu-Hu/action-persistence.git}

\noindent
\textbf{Author contributions}
Jianshu Hu contributed to the methodology, the experiments, and the paper writing.
Paul Weng and Yutong Ban contributed to the methodology and the paper writing.
\end{declaration}

\begin{appendices}

\section{Hyperparameters}\label{appendix:hyperparameters}
The commonly used hyperparameters across all experiments are shown in \Cref{tab:hyperparameters_DMC_drqv2}.
We just adopt the same hyperparameters from the DrQv2 \citep{DrQv2}.
For a fair comparison of using different action persistences, some hyperparameters are adjusted, as shown in \Cref{tab:different_hyperparameters_DMC}, to make sure the agent is trained with same number of updates under different settings. 
Moreover, humanoid tasks in DMControl are extremely hard.
So following the setting in original DrQv2, the hyperparameters such as total training frames are tuned in these tasks, shown in \Cref{tab:different_hyperparameters_DMC_humanoid}.

\begin{table}[ht]
\centering
\caption{Common hyperparameters used in experiments on DMControl (DrQv2)}
\label{tab:hyperparameters_DMC_drqv2}
{
\begin{tabular}{cc}
    \toprule
    Hyperparameter  & Value on DMC\\
    \midrule
    frame rendering & 84 $\times$ 84 $\times$ 3\\
    stacked frames  & 3\\
    replay buffer capacity & $10^6$\\
    seed frames & 4000\\
    exploration steps & 2000\\
    batch size $N$ & 256 \\
    discount $\gamma$ & 0.99\\
    optimizer ($\phi$, $\theta$)   & Adam\\
    learning rate ($\phi$,$\theta$) & 1e-4 \\
    agent update frequency & 2 \\
    target network soft-update rate & 0.01\\
    DDPG exploration stddev clip & 0.3\\
    DDPG exploration stddev schedule & linear(1.0, 0.1, 500000)\\
    epsilon-greedy exploration schedule & linear(1.0, 0.1, 500000)\\
    zeta in epsilon-zeta-greedy & 2 \\
    coefficient for calculating repeat probability & 1.0 \\
    \bottomrule
\end{tabular}}
\end{table}

\begin{table}[ht]
\centering
\caption{Different hyperparameters used in experiments on DMControl when using different action persistences}
\label{tab:different_hyperparameters_DMC}
{
\begin{tabular}{cccc}
    \toprule
    Hyperparameter  & Action persistence 1 & Action persistence 2 & Action persistence 4\\
    \midrule
    action persistence  & 1 & 2 & 4\\
    n-step returns & 6 & 3 & 2\\
    update every steps & 4 &2 &1\\
    \bottomrule
\end{tabular}}
\end{table}

\begin{table}[ht]
\centering
\caption{Different hyperparameters used in experiments on DMControl humanoid tasks}
\label{tab:different_hyperparameters_DMC_humanoid}
{
\begin{tabular}{cc}
    \toprule
    Hyperparameter  & humanoid tasks\\
    \midrule
    total training frames  & $10^7$\\
    DDPG exploration stddev schedule & linear(1.0, 0.1, 20000000)\\
    epsilon-greedy exploration schedule & linear(1.0, 0.1, 5000000)\\
    \bottomrule
\end{tabular}}
\end{table}

\begin{figure}[tb]
\centering
\begin{subfigure}[b]{0.32\textwidth}
\includegraphics[width=\linewidth]{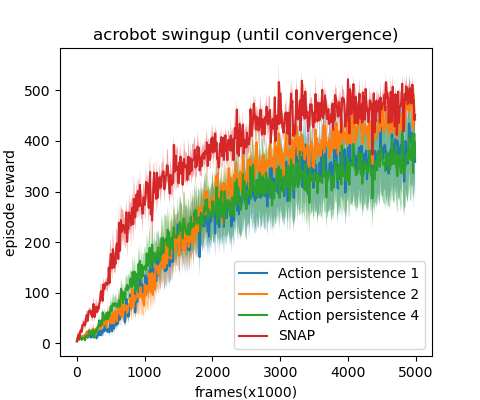}
\end{subfigure}
\begin{subfigure}[b]{0.32\textwidth}
\includegraphics[width=\linewidth]{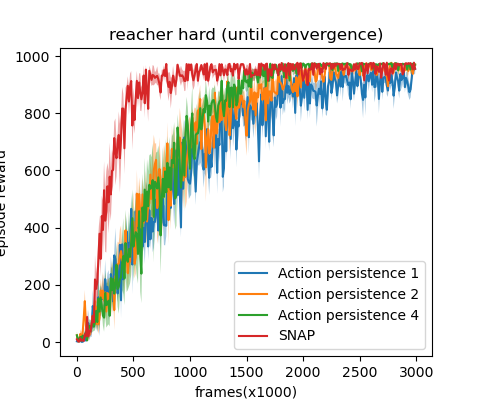}
\end{subfigure}
\begin{subfigure}[b]{0.32\textwidth}
\includegraphics[width=\linewidth]{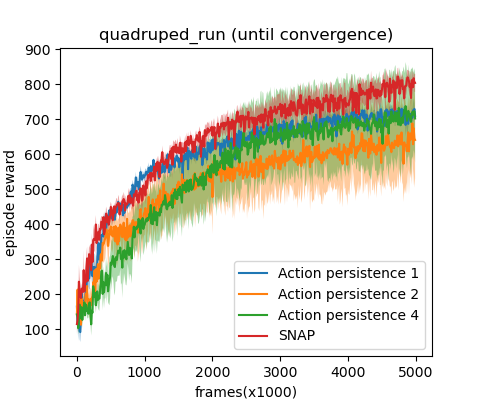}
\end{subfigure}
\caption{
{\textbf{DrQv2 with different action persistences.} The agents are trained until convergence for comparing the convergence rewards.
}}
\label{fig:convergence}
\end{figure}

{
\section{Additional results}
\label{appendix:additional_exps}
We also run experiments of using different action persistences $\ap={1,2,4}$ in Reacher-hard, Acrobot-swingup and Quadruped-run until convergence, as shown in \Cref{fig:convergence}.}




\end{appendices}


\bibliography{sn-bibliography}

\end{document}